\pdfoutput=1

\documentclass[11pt]{article}

\usepackage{EACL2023}

\usepackage{times}
\usepackage{latexsym}
\usepackage{courier}

\usepackage[T1]{fontenc}

\usepackage[utf8]{inputenc}

\usepackage{microtype}

\usepackage{inconsolata}
\usepackage{graphicx,tabularx,multirow,makecell}
\usepackage[section]{placeins}
\usepackage{float}

\newcolumntype{P}[1]{>{\centering\arraybackslash}p{#1}}
\newcolumntype{Y}{>{\centering\arraybackslash}X}

\title{Establishing degrees of closeness between audio recordings along different dimensions using large-scale cross-lingual models}

\author{Maxime Fily$^{1,2}$ \and Guillaume Wisniewski$^1$ \and Séverine Guillaume$^2$ \AND Gilles Adda$^3$ \and Alexis Michaud$^2$\\
\hfill \\
$^1$LLF, CNRS, Université Paris-Cité, F-75013, Paris, France\\
$^2$LACITO, CNRS, Université Sorbonne Nouvelle, F-94800, Villejuif, France\\
$^3$LISN, CNRS, Université Paris-Saclay, F-91405, Orsay, France
}

\begin{document}
\maketitle
\begin{abstract}
In the highly constrained context of low-resource language studies, we explore vector representations of speech from a pretrained model 
to determine their level of abstraction with regard to the audio signal. We propose a new unsupervised method using \texttt{ABX} tests on audio recordings with carefully curated metadata to shed light on the type of information present in the representations.
\texttt{ABX} tests determine whether the representations computed by a multilingual speech model encode a given characteristic. Three experiments are devised: one on room acoustics aspects, one on linguistic genre, and one on phonetic aspects.
The results confirm that the representations extracted from recordings with different linguistic/extra-linguistic characteristics differ along the same lines. 
Embedding more audio signal in one vector better discriminates extra-linguistic characteristics, whereas shorter snippets are better to distinguish segmental information. The method is fully unsupervised, potentially opening new research avenues for comparative work on under-documented languages.
\end{abstract}

\section{Introduction}

In recent improvements in speech processing,\footnote{In ASR, TTS, and even on corpora/languages/tasks not seen at pre-training \citep{guillaume2022computEL}.} the amount of data used at pre-training has been instrumental \citep{wei2022emergent}, which makes it more challenging~-- if not impossible~-- to reach similar levels of performance for endangered languages. Developing new unsupervised approaches, in addition to being cost-effective \citep{bender2021dangers}, helps us better understand speech models.


Speech 
is highly multifactorial: a recorded voice tells a message and conveys an intention, and the audio also contains information about the surroundings. This study addresses the topic of the nature of the information encoded in the representations produced by a neural network in an unsupervised manner. Towards this end, we perform distance measurements over the representations. Our goal is to investigate the level of abstraction encapsulated in these representations.

Our experimental setup relies on tailored datasets to see how specific differences in the input signal are reflected in the output vectors. \texttt{ABX} tests are used on audio data in the Na language (ISO-639-3: nru) and in the Naxi language (nxq). Three series of experiments are devised to assess differences between recordings. (i)~The \textit{folk tale series} aims to explore an extra-linguistic dimension by comparing seven versions of the same tale by the same speaker. (ii)~The \textit{song styles series} compares different songs interpreted by a single singer. (iii)~Finally, the \textit{phonetics series} explores the segmental dimension by comparing sentences (some identical, some different) from different speakers.

The results provide an insight into the nature of the information encoded in the representations of a model such as XLSR-53 \citep{baevski2020wav2vec,babu2021xls}. Our findings suggest that \texttt{ABX} tests can be leveraged to bring out differences in the acoustic setup (room, microphone), in the voice properties, or in the linguistic content. A parametric study shows that processing audio by snippets\footnote{The term `snippet' is preferred over `segment', reserving the latter to refer to phonetic segments.} of 10\,s is sufficient to bring out differences in the acoustic setup and in voice properties, while 1\,s snippets are better for segmental characteristics.

This study offers an innovative method to detect confounding factors in corpora intended for unsupervised learning, and provides a means to accelerate the classification of recordings (e.g., by noise level or genre) where such metadata are unavailable.

\section{Method \label{sec:method}}

We propose a method based on two components: (i)~\texttt{ABX} tests to determine -- via similarity tests -- whether a characteristic of an audio recording is present or not, and (ii)~audio corpora with precise metadata. These metadata allow us to build datasets based on one characteristic at a time: language name, speaker ID, room acoustics, microphone type, voice properties or segmental content.

\paragraph{ABX tests} To find out, in an unsupervised manner, if a multilingual speech model encodes a characteristic $\mathcal{C}$ of the speech signal, we use the \texttt{ABX} tests introduced by \newcite{carlin2011rapid} and \newcite{schatz2013evaluating}. The test relies on vector representations built by a pre-trained model for three audio snippets. Let $A$ and $X$ denote the snippets that share the characteristic $\mathcal{C}$, while $B$ is the one that does not. The test checks whether the distance $d(A, X)$ is smaller than $d(A, B)$. The metric used in our \texttt{ABX} tests is the cosine distance.

The \texttt{ABX} score corresponds to the proportion of triplets for which the condition $d(A, X)~<~d(A, B)$ holds true. An \texttt{ABX} score close to 50\,\% (or lower) indicates that, on average, the distance between $A$ and $X$  is close to the distance between $A$ and $B$, suggesting that  $\mathcal{C}$ is not encoded in the audio representation. Conversely, the closer the score is to 100\,\%, the more the representation captures the characteristic $\mathcal{C}$.

\texttt{ABX} tests are interesting for low-resource scenarios because they require no additional training, so they can be directly applied to the representations (unlike linguistic probes:  \citealt{belinkov2019analysis,belinkov2017analyzing,yin2022interpreting}).


\paragraph{Corpora} Our study relies on recordings in Na (ISO-639-3 code: nru) and Naxi (nxq). Na and Naxi are spoken in Southwest China. Na is the mother tongue of approximately 50,000 people. Naxi is more widely spoken, as the mother tongue of approximately 200,000 people. Both languages are gradually replaced by Mandarin, the official language used in schools, administrations and the media \cite{michaudetal2011,zhao2022looking}.
All recordings come from the \href{https://pangloss.cnrs.fr}{Pangloss} Collection, an open-access archive of ‘little-documented languages’. Each resource's \textsc{doi} is provided in App.~\ref{sec:appendix3}. Three series of recordings selected for their characteristics are considered:

(i) The \textbf{\textit{folk tale series}} consists of seven recording sessions of the same folk tale in Na, told by the same speaker. These experiments focus on the effect of the recording conditions, which are slightly different from one version to another, and for which \texttt{ABX} tests are performed. For example, $V_1$ (A) is compared to $V_3$ (B), and for that we assume that $V_1$ is X and calculate $d(V_1,V_1)$ vs $d(V_1,V_3)$. If $d(V_1,V_3) > d(V_1,V_1)$ \textit{more often} than $d(V_1,V_3) < d(V_1,V_1)$, then we assume that $V_1$ and $V_3$ are distinguished.

The first batch studied comprises three versions: $V_1$, $V_2$ and $V_3$. $V_1$ was recorded in a room with perceptible reverberation, while $V_2$ and $V_3$ were recorded in a damped room. 

The second batch is made up of $V_6$ and $V_7$. These two versions were recorded in the same acoustic conditions. The audio was captured simultaneously by two microphones: a headset microphone and a handheld microphone placed on a small stand.

The third batch compares $V_4$ and $V_5$ to all the other recordings of the \textit{folk tale series}. $V_4$ and $V_5$ have a native listener acting as respondent.

These recordings are particularly interesting because some potential confounding factors (typically the topic and the speaker) are controlled, which makes it possible to focus on the influence of certain specific factors (e.g.,\ room acoustics).

(ii) The \textbf{\textit{song styles series}} consists of five recordings of the same Naxi professional singer. Three only-song recordings are considered, one narrative and one recording with both genres (``Alili'', 50\,\% text, 50\,\% song). The aim is to compare these recordings. A trained singer exhibits very different voice properties when singing and talking. Vowel quality and tessitura are affected \citep[][458]{furniss2016michele}. Such differences are perceptible and categorized differently by listeners \citep[][187]{furniss2016michele}. This experiment aims to check if this is reflected in the representations.

(iii)~The \textbf{\textit{phonetics series}} is made up of five recordings of phonetic elicitations and one recording of words in a carrier sentence, in the Na language. Three speakers identified as AS, RS and TLT are considered. We included two recording sessions, which allows for intra-speaker comparison.

The five recordings of phonetic elicitations have the same content (apart from the variation inherent to the experimental process in fieldwork conditions: \citealt{niebuhr2015speech}) whereas lexical elicitations are a completely different content. Only AS participated in both the phonetic and lexical elicitation sessions.

Tables~\ref{tab:tale_metadata}, \ref{tab:song_metadata} and~\ref{tab:phonet_metadata} in App.~\ref{sec:appendix0} provide a more complete view of the abovementioned metadata.


\paragraph{Experimental Setting} In all our experiments, we use the \texttt{XLSR-53}\footnote{The HuggingFace API was used (model signature: \texttt{facebook/wav2vec2-large-xlsr-53}).} model, a \texttt{wav2vec2} architecture trained on 56\,kh of (raw) audio data in 53~languages \cite{conneau20unsupervised}. Na is not present in the pre-training data of this model, but it has been shown that the model can be fine-tuned to do ASR on Na \citep{guillaume2022computEL}, and therefore the phonetic module is able to handle the diversity of surface realizations of this language.
For the comparisons, we consider audio snippets of length 1\,s, 5\,s, 10\,s and 20\,s in order to study the effect of snippet length on our \texttt{ABX} test. We use max-pooling to build a single vector representing the snippet. We then build fine-grained heatmaps of \texttt{ABX} scores.

We use the representations from the 21\textsuperscript{st} layer, following tests on a validation set. This choice is based on the findings of \citet{pasad2021layer, pasad2023comparative} and \citet{li2022fusing, li2023exploration}, who show that the ability of \texttt{wav2vec2} representations to capture linguistic information declines in the final three layers.

\section{Results \label{res}}

Using \texttt{ABX} tests with carefully selected audio recordings, we investigate whether or not the audio representations computed by \texttt{wav2vec2} capture specific information from the audio signal.

\subsection{Study of various versions of the same tale}

The aim of this experiment is to determine whether certain extra-linguistic variables (e.g.,\ room acoustics, and type of microphone) are captured in the neural representations. For that, we consider recordings from the \textit{folk tale series} and use \texttt{ABX} tests to distinguish between different versions of the tale: these scores are calculated from triplets consisting of two snippets of 10\,s from the same version and one snippet from a different version.\footnote{Results for other snippet lengths are reported in App.~\ref{sec:appendix1}.}

\begin{figure}[hbtp]
\includegraphics[width=0.49\textwidth]{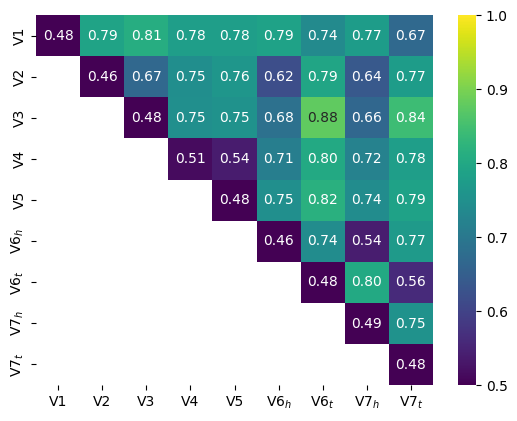}
    \caption{\texttt{ABX} scores when distinguishing different versions of the \textit{folk tale series}. Snippet length = 10\,s.}
    \label{fig:sister_tri_10s_body}
    \vspace{-4mm}
\end{figure}

Figure~\ref{fig:sister_tri_10s_body} shows that, in most cases, with a 10\,s snippet-length it is possible to distinguish between the different recordings, although it is always the same speaker telling the same story: except for a few rare exceptions, which are addressed later, most of the reported scores are well above 50\,\%. What is more, the scores on the diagonals, corresponding to tests where all the excerpts come from the same recording, are all close to 50\,\%. This clearly indicates that the differences found in the other \texttt{ABX} tests are not due to linguistic content (the words spoken), but rather to acoustic configuration. It suggests that neural representations capture much more than the linguistic information needed to understand speech, and it seems possible to use them to retrieve information related to the recording conditions. 

A more precise analysis of the scores between two recording conditions provides a better understanding of the  information that is or is not captured by the representations.

\begin{figure}[H]
     \centering
     \includegraphics[width=0.32\textwidth]{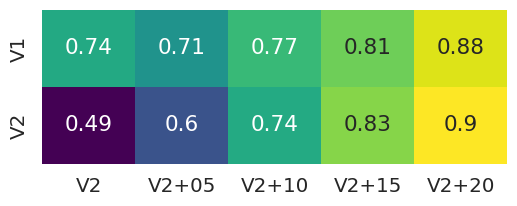}
     \caption{Reproducing $V_1$ room tone with artificial room tone applied on $V_2$. Snippet length = 5\,s.}
     \label{fig:artif_roomtone}
\end{figure}

The first batch is a comparison between $V_1$, $V_2$ and $V_3$ (NW corner of Figure~\ref{fig:sister_tri_10s_body}): the \texttt{ABX} scores show that the representation of $V_2$ and $V_3$ are indistinguishable when compared to the representations of $V_1$ (0.79 vs 0.81). We know from Section~\ref{sec:method} that the main difference between these three recordings is related to the recording venue: $V_2$ and $V_3$ were recorded in the same place, less reverberating than the place where $V_1$ was recorded. To confirm the influence of this parameter, we carried out a complementary experiment by artificially adding \textit{reverb}\footnote{We use Audacity to add 5, 10, 15 or 20\,\% reverb.} to the $V_2$ recordings and measuring the \texttt{ABX} score between the $V_1$ and modified $V_2$ recordings. Figure~\ref{fig:artif_roomtone} shows the evolution of the \texttt{ABX} score as a function of the amount of reverb added. One interesting observation is that when gradually increasing the amount of reverb in $V_2$, the \texttt{ABX} score decreases first before increasing again. It means that $V_1$ is closer to $V_2$ with 5\,\% reverb, which suggests a relation of causality between the amount of reverberation and the degree of closeness between the recordings of this batch.


In the second batch, the sub-versions of $V_6$ and $V_7$ are labeled as $h$ for \textit{headset} and $t$ for \textit{table} (remember that the two types of microphone used are (i)~headset microphone and (ii)~handheld microphone placed on a small stand, on a table). Figure~\ref{fig:sister_tri_10s_body} shows that the \texttt{XLSR-53} representations can effectively distinguish between microphone types with high precision. For instance, the \texttt{ABX} scores between V$_{6, h}$ and V$_{6, t}$ are some of the highest in our experiment. However, when it comes to distinguishing between two different recordings made with the same microphone (i.e.\, V$_{6, h}$-V$_{7, h}$ and V$_{6, t}$-V$_{7, t}$), the \texttt{ABX} scores are only slightly better than scores for the same recording. This suggests that the representations, extracted in 10\,s~long snippets, strongly depend on the microphone used: two vectors representing the same audio signal but recorded by different microphones come out as more dissimilar than those representing two different audio signals recorded by the same microphone.

Figure~\ref{fig:sister_tri_10s_body} also brings out uncanny similarity between 
recordings $V_4$ and $V_5$. 
The \texttt{ABX} score between these is only 54\,\%, whereas it is no lower than 71\,\% for all other pairs. Now, $V_4$ and $V_5$ are the only recordings at which a listener from the language community was present: the others were produced with just the investigator~-- who has low fluency in Na~-- as audience. This looks like a case of linguistic adaptation \citep{piazza2022acoustic}. It suggests possibilities for automatically generating hypotheses about the communicative setting of a recording. 


In this experiment series, all our observations are most visible with 10\,s snippets, which seems to be the proper setting to reveal differences at a broad acoustic level. It also seems to be a suitable snippet size to reveal differences at the prosodic level. Further experiments are necessary to confirm our conclusions.

\subsection{Study of different song styles}

The aim of this experiment is to explore whether or not the extraction settings devised in the preceding experiment allow us to explore the representations with regard to the voice properties of the speaker. Several recordings of a professional Naxi singer are compared to one another : one song in the ``Alili'' style, two in the ``Guqi'' style, one in the ``Wo~Menda'' style, and one narrative. The songs originally contained a non-sung introduction which has been removed for the comparisons, except for the ``Alili''-style song, which is half-text and half-song.

\begin{figure}[hbtp]
    \includegraphics[width=0.49\textwidth]{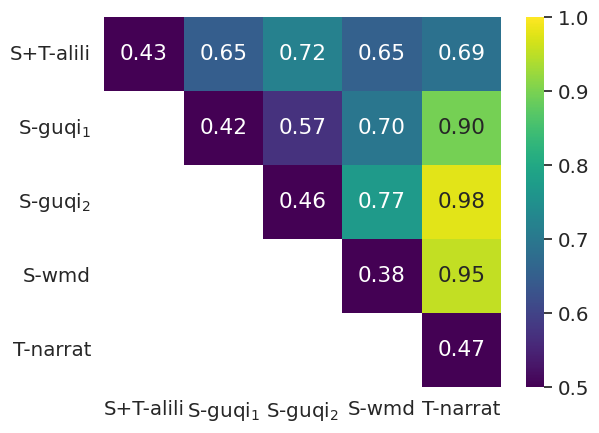}
    \caption{\texttt{ABX} scores for the comparisons between different genres (T=text (narrative), S=song). Songs in three different styles and narratives are performed by a professional Naxi singer. Snippet length = 10\,s.}
    \label{fig:songtri_10s}
    \vspace{-2mm}
\end{figure}

Figure~\ref{fig:songtri_10s} shows that all the songs are strongly distinguished from the narrative, except for the ``Alili'' recording, which is half-text half-song. Interestingly, the ``Alili'' recording patterns neither with the songs nor with the narrative: it stands halfway between. As for the two songs in the ``Guqi'' style, they exhibit the lowest \texttt{ABX} score (0.57), which suggests that song style may be detectable. 

These results suggest that voice properties are present in the representations, since we can distinguish between a narrative and various song styles for the same speaker, and even regroup by song style. These results are very encouraging for future studies that aim at using neural models to perform prosodic studies.

\subsection{Study of a phonetics corpus}

While it is quite obvious that two sentences with a different linguistic content in perfectly controlled conditions will come out as different when submitted to an \texttt{ABX} test, the answer is not immediate when it comes to a whole recording. It is also not obvious that two different sentences uttered by two different speakers are distinguished solely due to a difference in the linguistic content: speaker ID acts as a confounding factor.

The aim of this experiment is to perform \texttt{ABX} tests on data with differences on the phonetic segments. To do this, we rely on a phonetics corpus recorded in a controlled manner, where each speaker received similar instructions. Some recordings have the same content (AS$_{1,2}$, RS$_{1,2}$, TLT), and one recording has a different content (AS$_{lex}$). The scores are calculated from triplets consisting of two snippets of 1\,s from the same recording and one snippet from a different recording.\footnote{Results for other snippet lengths are reported in App.~\ref{sec:appendix2}.}

\begin{figure}[hbtp]
    \includegraphics[width=0.49\textwidth]{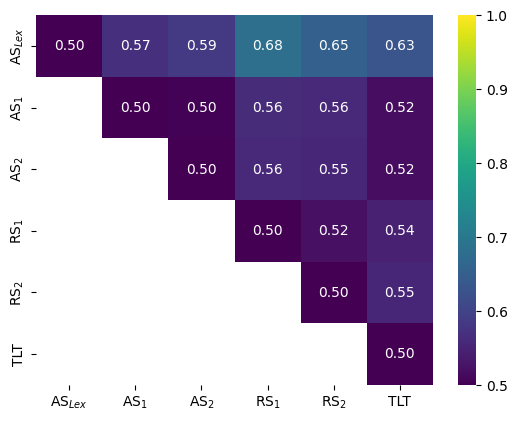}
    \caption{\texttt{ABX} scores for comparisons within 
    the \textit{phonetics series}. Speaker AS has three recordings (AS$_1$, AS$_2$, AS$_{Lex}$), RS has two (RS$_1$, RS$_2$) and TLT has one. Snippet length = 1\,s.}
    \label{fig:phontri_1s_body}
    \vspace{-2mm}
\end{figure}

First, Figure~\ref{fig:phontri_1s_body} shows that with a 1\,s snippet-length it is nearly not possible to distinguish between the different recordings of the same sentences, even when the speakers differ. It suggests that neural representations, in this configuration, effectively `centrifugate' the extra-linguistic information. This observation is not surprising given how the models are pre-trained \citep{baevski20}, and it is a convenient springboard for the second part of the analysis, which consists in comparing these recordings of identical sentences to another one with different sentences.

The results in the first row of Figure~\ref{fig:phontri_1s_body} indeed suggest that the \texttt{ABX} tests reveal differences in linguistic content. The magnitude of the discrepancy (between row 1 and the others) depends on whether or not the speaker is different. The fixed-speaker discrepancy is around 0.07, while the cross-speaker discrepancy is around 0.11. It suggests that even with 1\,s snippets, speaker ID is still reflected in some way in the representations.

In this study, \texttt{ABX} scores are averaged over an entire recording. For phonetic differences, it would be interesting to be able to perform comparisons on a per-sentence basis, but it would constitute a departure from a fully unsupervised approach.



\section{Discussion and conclusion}

When one undertakes the task of comparing vector representations of audio, differences are expected, too many of them rather than too few. We adopted an experimental method to submit a given model to different experiments with test variables. 

In the first two series, the recordings are distinguished according to (i) technical acoustic properties in the \textit{folk tale series}, or (ii) voice properties in batch $V_4$, $V_5$ of the \textit{folk tale series} or in the \textit{song styles series}. A 10\,s snippet length seems to best reveal differences in characteristics such as (i) room acoustics or microphone type or (ii) speech rate or genre. Our aim in these two series was to explore to what degree extra-linguistic information is present in the representations. Being able to detect acoustic differences such as the amount of reverb in a room, or the fact that we are not only capable of measuring differences between narratives and songs but also to distinguish between song styles, gives us reasons to think that our method should be useful to automatically classify recordings based on room acoustics, interview setup, or genre. The prosodic characteristics of a recording also seem to be encoded, which is encouraging for future research on tone using unsupervised methods on audio recordings.

In the \textit{phonetics series}, we focused on 1\,s snippet lengths. The recordings of three speakers who participated in a phonetics experiment, quasi-identical to one another, are distinguished from a recording with a different content, but the distinction is not very strong. The snippets from this series are shorter and result in smaller differences on the \texttt{ABX} score. This observation suggests that differences are only detected when the segmental content changes, and shows the consistency of our method. Using this method on cross-speaker, or cross-linguistic snippets however requires additional investigations to devise a method more suited to phonetic segments. Among possible improvements, using segmented corpora would be an interesting avenue of research. 





\section*{Limitations}

 As is often the case for endangered languages \citep{liu2022investigating}, our corpora rely on a few speakers of the same gender. In our case, we exploit a resource with rich metadata to build experiments with minimal differences and observe sets that differ by one characteristic only. The conclusions drawn on the speaker-independent setting in Section~\ref{res} may need to be reanalyzed when we run the experiment on cross-gender data.
 
Our study does not perform comparisons with other methods for identifying characteristics, because other methods require more data than the amount treated here (typically linguistic probes using classifiers).

We have not investigated how the model reacts to a superposition of variables sensitive to a given snippet length. Therefore, we would need to extend  our experiments further, e.g., to check how a 10~s snippet length is handled when assessing a discrepancy in speaker and room acoustics.

We plan to extend this study by adding data from experimental phonetics experiments related to second language acquisition, as they often include productions from the same speaker in multiple languages. Experimental phonetics corpora are devised under highly controlled conditions, which is beneficial for our study as it removes potential confounding factors.
 



\section*{Ethics Statement}
The study presented here relies on small-sized corpora because the methods are meant for low-resource languages, i.e., without a significant amount of data available. This limitation is offset by the wealth of metadata available for each recording in the Pangloss Collection. Pangloss is a world language open-access archive developed in a Dublin-core compliant framework \citep{weibel1998dublin}.

The data used in this study are first-hand, collected by researchers working with the communities to document and describe their language. They are the result of months of collaborative work in the field to transcribe and translate the data with native speakers (typically the speaker himself/herself). The speakers all consented to the use of these data for scientific purposes and were compensated for their work as linguistic consultants.

 All data and models in this study are open-access under a Creative Commons license stated on the consultation page for each resource (which is also the landing page of its \textsc{doi} listed in Table~\ref{tab:doi}). The information needed for reproducibility is present in the text (model information) or the appendices (data). The metadata collected were directly collected via questionnaires during the fieldwork. Gender, for example, corresponds to the gender the speaker provided in the questionnaire.

\section*{Acknowledgments}

We are grateful for constant support from the Na and Naxi communities. We would like to especially thank 
Mrs.\ Wang Sada and Mrs.\ Latami Dashilame of the Na community for their sharing of their expertise, their generosity with their time, and their confidence and encouragement. 

This research was partially funded by the \textsc{DiagnoSTIC} project  supported by the \textit{Agence d'Innovation de Défense} (grant n\textsuperscript{o} 2022 65 007) and the \textsc{DeepTypo} project supported by the \textit{Agence Nationale de la Recherche} (ANR-23-CE38-0003-01).


We are grateful to the sponsors of the field trips that made data collection on Naxi and Na possible (from 2002 to 2019). Specifically, we wish to acknowledge the Grenoble UGA IDEX international mobility program's support for fieldwork on Lataddi (Shekua) Na in 2019.





\bibliography{anthology,custom,refs}
\bibliographystyle{acl_natbib}

\appendix
\clearpage

\section{Metadata for the experiments \label{sec:appendix0}}
The list of metadata for the experiments conducted is given in Table~\ref{tab:tale_metadata} for the \textit{folk tale series}, Table~\ref{tab:song_metadata} for the \textit{song styles series} and in Table~\ref{tab:phonet_metadata} for the \textit{phonetics series}.
\begin{table}[ht]
\begin{center}
\begin{tabular}{cccccc}
\hline
\textsc{rec id}&Year & \textsc{dur} (s) & \textsc{mic} & \textsc{itv}&Acoust.\\ 
\hline
V1 & 2006 & 518  & Tab & out & ND \\ 
V2 & 2007 & 440  & Tab & out & D \\ 
V3 & 2008 & 707  & Tab & out & D \\ 
V4 & 2014 & 527  & Hea & Na & D \\ 
V5 & 2014 & 423  & Hea & Na & D \\ 
V6$_h$ & 2018 & 348  & Hea & out & ND \\ 
V6$_t$ & 2018 & 348  & Tab & out & ND \\ 
V7$_h$ & 2018 & 635  & Hea & out & ND \\ 
V7$_t$ & 2018 & 635  & Tab & out & ND \\ 
\hline
\end{tabular}
\caption{Metadata for the \textit{folk tale} series.  \textsc{mic} = microphone: Headset or Table; \textsc{itv} = interviewer: outsider or Na (local). Acoustics: non-damped (ND), or damped (D).}
\label{tab:tale_metadata}
\end{center}
\vspace{-2mm}
\end{table}

\begin{table}[ht]
\begin{center}
\begin{tabular}{lrr}
\hline
\multicolumn{1}{c}{\textsc{rec id}} & \multicolumn{1}{c}{\textsc{dur} (s)} & \multicolumn{1}{c}{\textsc{\%\,song}}\\ 
\hline
S-guqi$_1$ & 151 & 100 \\ 
S-guqi$_2$ & 300 & 100 \\ 
T-narrat & 296 & 0 \\ 
S-wmd & 129 & 100 \\ 
S+T-alili & 194 & 49 \\ 
\hline
\end{tabular}
\caption{Metadata for the \textit{song styles series}, including the ratio of sung voice over recording duration.}
\label{tab:song_metadata}
\end{center}
\vspace{-4mm}
\end{table}

\begin{table}[ht]
\begin{center}
\begin{tabular}{lrll}
\hline
\multicolumn{1}{c}{\textsc{rec id}} & \multicolumn{1}{c}{\textsc{dur} (s)} & \multicolumn{1}{c}{\textsc{spk}} & \multicolumn{1}{c}{\textsc{session type}}\\ 
\hline
AS$_1$ & 1567 & AS (F) & Phonetic elicit.\\ 
AS$_2$ & 952 & AS (F) & Phonetic elicit.\\ 
RS$_1$ & 681 & RS (F) & Phonetic elicit.\\ 
RS$_2$ & 786 & RS (F) & Phonetic elicit.\\ 
TLT & 897 & TLT (F) & Phonetic elicit.\\ 
AS$_{Lex}$ & 1216 & AS (F) & Lexical elicit.\\ 
\hline
\end{tabular}
\caption{Metadata for the \textit{phonetics series}. \textsc{spk} = speaker; (F) = Female. Data collected in 2019}
\label{tab:phonet_metadata}
\end{center}
\vspace{-4mm}
\end{table}

\section{M and SD values showing that \texttt{ABX} tests can be used to measure differences between our corpora}

\begin{figure}[H]
    \centering
    \includegraphics[width=0.5\textwidth]{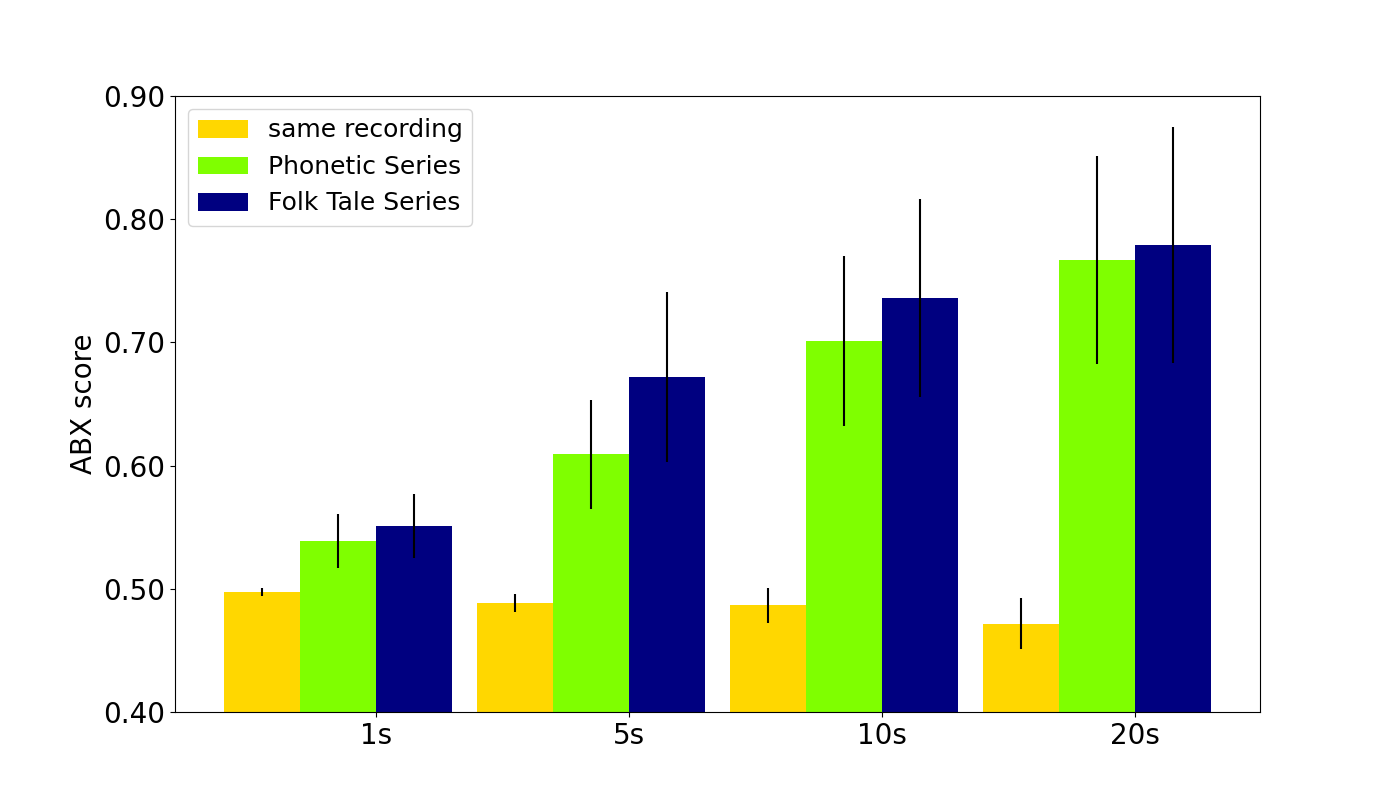}
    \caption{Average \texttt{ABX} scores for 1, 5, 10, 20\,s snippets.}
    \label{fig:errorevol}
    \vspace{-2mm}
\end{figure}

Figure~\ref{fig:errorevol} shows mean and standard deviation values for a comparison between inter-recordings scores (\textit{phonetics series} and \textit{folk tale series} barplots) and intra-recording scores (\textit{same-recording}), for different snippet lengths. For all snippet lengths, the average inter-recording \texttt{ABX} score is always significantly higher than the average intra-recording score, even for 1\,s snippet-length. This shows that \texttt{ABX} tests can be used to measure differences in our experiments.


\section{ABX scores when distinguishing different versions of the \textit{folk tale series} by the same speaker.\label{sec:appendix1}}

\begin{figure}[H]
\includegraphics[width=0.49\textwidth]{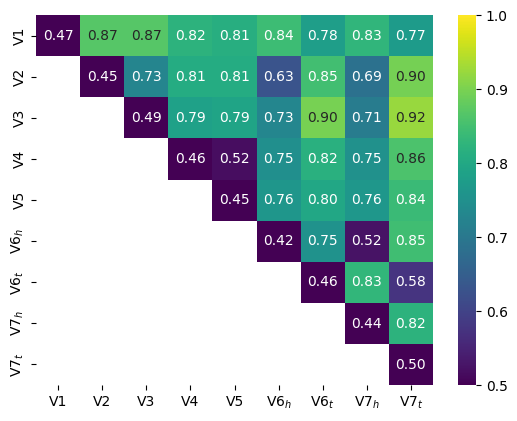}
    \caption{\texttt{ABX} scores for the \textit{folk tale series}. (snippet length~=~20\,s).}
    \label{fig:sister_tri_20s}
\end{figure}

\begin{figure}[H]
\includegraphics[width=0.49\textwidth]{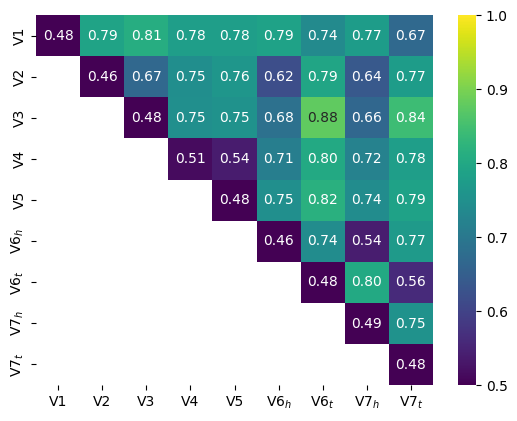}
    \caption{\texttt{ABX} scores for the \textit{folk tale series} (snippet length~=~10\,s).}
    \label{fig:sister_tri_10s}
\end{figure}

\begin{figure}[H]
\includegraphics[width=0.49\textwidth]{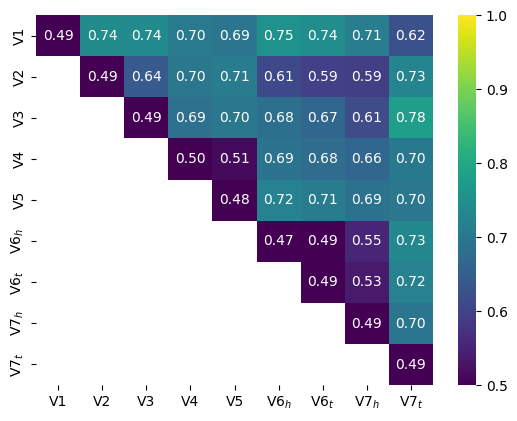}
    \caption{\texttt{ABX} scores for the \textit{folk tale series} (snippet length~=~5\,s).}
    \label{fig:sister_tri_5s}
\end{figure}

\begin{figure}[H]
\includegraphics[width=0.49\textwidth]{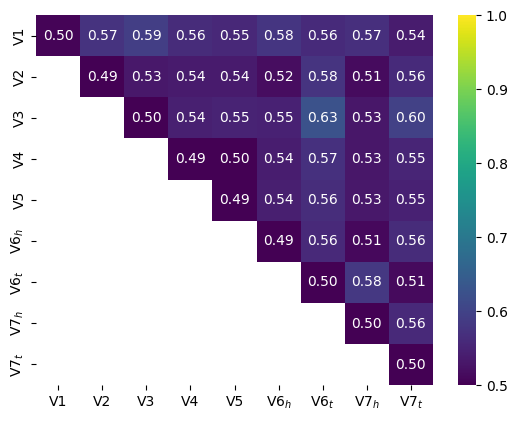}
    \caption{\texttt{ABX} scores for the \textit{folk tale series} (snippet length~=~1\,s).}
    \label{fig:sister_tri_1s}
\end{figure}

The 20~s value for snippet length has been investigated, and it does not bring out much more than the 10~s snippet length. In addition a 20\,s snippet length with max-pooling tackles the limits of the max-pooling method. Indeed, we believe there is a limit to the amount of audio we can have in an embedding. Indeed, with the max pooling extraction method, each of the 980 vectors before pooling the 20~s of audio will only occupy, on average, 1.04 cells per final vector since it only has 1,024 components. The results can be seen in Figure \ref{fig:sister_tri_20s} for 20\,s snippets, Figure \ref{fig:sister_tri_10s} for 10\,s snippets, Figure \ref{fig:sister_tri_5s} for 5\,s snippets, Figure \ref{fig:sister_tri_1s} for 1\,s snippets.

\section{\texttt{ABX} scores when distinguishing between elements of the \textit{phonetics series}}
\label{sec:appendix2}

\begin{figure}[hbtp]
    \includegraphics[width=0.49\textwidth]{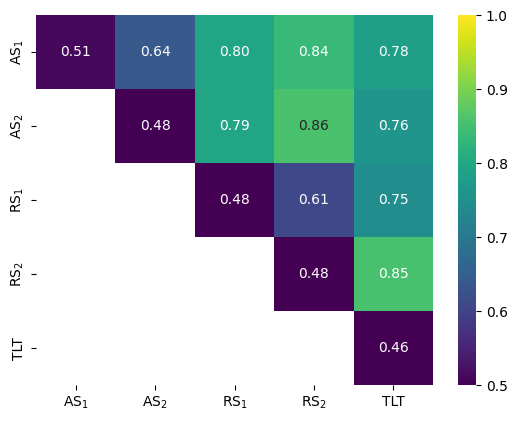}
    \caption{\texttt{ABX} scores for the comparisons between elements of the \textit{phonetics series} (snippet length~=~20\,s). }
    \label{fig:phontri_20s}
\end{figure}

\begin{figure}[hbtp]
    \includegraphics[width=0.49\textwidth]{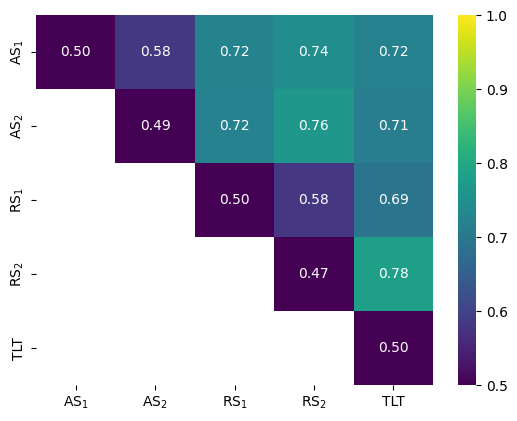}
    \caption{\texttt{ABX} scores for the comparisons between elements of the \textit{phonetics series} (snippet length~=~10\,s). }
    \label{fig:phontri_10s}
\end{figure}

\begin{figure}[hbtp]
    \includegraphics[width=0.49\textwidth]{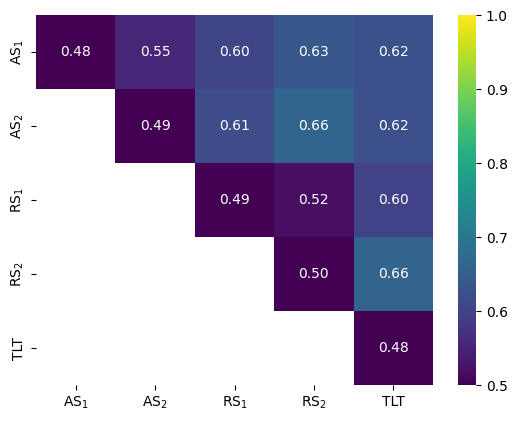}
    \caption{\texttt{ABX} scores for the comparisons between elements of the \textit{phonetics series}  (snippet length~=~5\,s). }
    \label{fig:phontri_5s}
\end{figure}

\begin{figure}[htp]
    \includegraphics[width=0.49\textwidth]{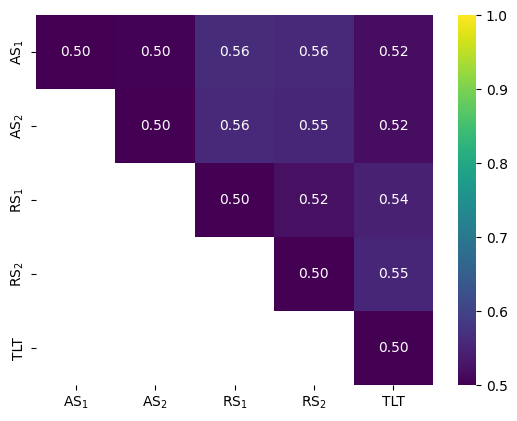}
    \caption{\texttt{ABX} scores for the comparisons between elements of the \textit{phonetics series} (snippet length~=~1\,s). }
    \label{fig:phontri_1s}
\end{figure}

The results can be seen in Figure \ref{fig:phontri_20s} for 20\,s snippets, Figure \ref{fig:phontri_10s} for 10\,s snippets, Figure \ref{fig:phontri_5s} for 5\,s snippets, Figure \ref{fig:phontri_1s} for 1\,s snippets.

\section{Audio resource: list of the recordings used for the study, with their \textsc{doi}}
\label{sec:appendix3}

\begin{table}[htbp]
\begin{tabular}{ll}
\hline
\multicolumn{2}{l}{\textit{\textbf{Folk tale series}}:} \\ 
\textsc{rec id} & \textsc{doi} \\ 
\hline
V1 & \small \url{doi.org/10.24397/PANGLOSS-0004341} \\ 
V2 & \small \url{doi.org/10.24397/PANGLOSS-0004343} \\ 
V3 & \small \url{doi.org/10.24397/PANGLOSS-0004344} \\ 
V4 & \small \url{doi.org/10.24397/pangloss-0004938} \\ 
V5 & \small \url{doi.org/10.24397/pangloss-0004940} \\ 
V6 & \small \url{doi.org/10.24397/pangloss-0007695} \\ 
V7 & \small \url{doi.org/10.24397/pangloss-0007698} \\ 
\hline
&\\
\hline
\multicolumn{2}{l}{\textit{\textbf{Song styles series}}:} \\ 
\textsc{rec id} & \textsc{doi} \\ 
\hline
S-guqi$_1$  & \small \url{doi.org/10.24397/pangloss-0004694} \\ 
S-guqi$_2$ & \small \url{doi.org/10.24397/pangloss-0004697} \\ 
T-narrat & \small \url{doi.org/10.24397/pangloss-0004695} \\ 
S-wmd & \small \url{doi.org/10.24397/pangloss-0004698} \\
S+T-alili & \small \url{doi.org/10.24397/pangloss-0004699} \\
\hline
&\\
\hline
\multicolumn{2}{l}{\textbf{Phonetics series}}  \\ 
\textsc{rec id} & \textsc{doi} \\ 
\hline
AS$_2$  &  \small \url{doi.org/10.24397/pangloss-0008663}\\ 
RS$_2$ & \small \url{doi.org/10.24397/pangloss-0008667} \\ 
AS$_1$  & \small \url{doi.org/10.24397/pangloss-0008662} \\ 
\multirow{3}{*}{RS$_1$} & \small \url{doi.org/10.24397/pangloss-0008664 }\\ 
& \small \url{doi.org/10.24397/pangloss-0008665} \\ 
& \small \url{doi.org/10.24397/pangloss-0008666} \\ 
TLT  & \small \url{doi.org/10.24397/pangloss-0008668} \\ 
\multirow{3}{*}{AS$_{Lex}$}  & \small \url{doi.org/10.24397/pangloss-0008669} \\ 
 & \small \url{doi.org/10.24397/pangloss-0008670}\\ 
 & \small \url{doi.org/10.24397/pangloss-0008671} \\ 
 \hline
\end{tabular}
\caption{List of the \textsc{doi}s for the recordings used in this study.}
\label{tab:doi}
\end{table}

\maketitle


\end{document}